\documentclass{article}
\usepackage{graphicx}
\graphicspath{ {./} }
\usepackage{amsmath}
\usepackage[numbers]{natbib}
    \usepackage[final]{tackling_climate_workshop_style}

\usepackage[utf8]{inputenc} % allow utf-8 input
\usepackage[T1]{fontenc}    % use 8-bit T1 fonts
\usepackage{hyperref}       % hyperlinks
\usepackage{url}            % simple URL typesetting
\usepackage{booktabs}       % professional-quality tables
\usepackage{amsfonts}       % blackboard math symbols
\usepackage{nicefrac}       % compact symbols for 1/2, etc.
\usepackage{microtype}      % microtypography

\title{Performance evaluation of deep segmentation models for Contrails detection}

\author{
  Akshat Bhandari\thanks{Manipal Institute of Technology, Manipal Academy of Higher Education, Manipal, India}\protect\phantom{\footnotesize 1}\thanks{Authors have contributed equally to this work}\\
  Dept. of Computer Science and Engineering \\
  \texttt{akshatbhandari15@gmail.com} \\
  \And
  Sriya Rallabandi\footnotemark[1]\protect\phantom{\footnotesize 1}\footnotemark[2]\\
  Dept. of Information and\cr Communication Technology \\
  \texttt{sriyarallabandi@gmail.com} \\
  \And
  Sanchit Singhal\footnotemark[1]\protect\phantom{\footnotesize 1}\footnotemark[2]\\
  Dept. of Electrical and Electronics Engineering \\
  \texttt{sanchitsinghal57@gmail.com} \\
  \And
  Aditya Kasliwal\footnotemark[1]\protect\phantom{\footnotesize 1}\footnotemark[2]\\
  Dept. of Data Science and\cr Computer Applications \\
  \texttt{kasliwaladitya17@gmail.com} \\
  \And
  Pratinav Seth\footnotemark[1]\\
  Dept. of Data Science and Computer Applications \\
  \texttt{seth.pratinav@gmail.com}\\
}

\begin{document}

\maketitle
\begin{abstract}
Contrails, short for condensation trails, are line-shaped ice clouds produced by aircraft engine exhaust when they fly through cold and humid air. They generate a greenhouse effect by absorbing or directing back to Earth approximately 33\% of emitted outgoing longwave radiation\cite{AVILA2019100033}. They account for over half of the climate change resulting from aviation activities. Avoiding contrails and adjusting flight routes could be an inexpensive and effective way to reduce their impact. An accurate, automated, and reliable detection algorithm is required to develop and evaluate contrail avoidance strategies. Advancement in contrail detection has been severely limited due to several factors, primarily due to a lack of quality-labeled data. Recently, \cite{mccloskeyhuman} proposed a large human-labeled Landsat-8 contrails dataset. Each contrail is carefully labeled with various inputs in various scenes of Landsat-8 satellite imagery. In this work, we benchmark several popular segmentation models with combinations of different loss functions and encoder backbones. This work is the first to apply state-of-the-art segmentation techniques to detect contrails in low-orbit satellite imagery. Our work can also be used as an open benchmark for contrail segmentation and is available at  \url{https://github.com/Kasliwal17/Contrail_Segmentation}.
\end{abstract}

\section{Introduction}
Contrails, or vapor trails, usually form when water vapor from the exhaust combines with the low ambient temperatures in high-altitude regions. These contrails create an additional blanket of clouds, increasing the heat-trapping effect that directly contributes to global warming. Although some may be short-lived, they can, in many cases, linger for hours\cite{vazquez2015contrail}, which results in new contrails created by aircraft flying on the same routes, accumulating and forming a 'contrail cirrus'\cite{bock2016reassessing}\cite{burkhardt2011global}, which worsens the heat-trapping effect. As a result, contrails have become a new cause of alarm for climate change as they have become one of the most significant contributors to global warming caused by the aviation industry\cite{pearce2019airplane}.

An effective solution that can reduce the effect of contrails is to avoid the regions of the atmosphere that are cold and humid enough to create contrails\cite{molloy2022design} and to adjust flight routes and slight changes in flight altitudes\cite{mannstein2005note} to reduce the air traffic in such regions. This can be a relatively easy solution to prevent contrail formations and their effects, given that a comparably small proportion of aircraft contributes majorly. Physics-based models such as CoCIP\cite{schumann2012contrail} and APCEMM\cite{fritz2020role} simulate contrails' microphysics and radiative transfer effects but suffer from uncertainties in their inputs. These uncertainties -such as the spatial resolution of numerical weather models and limitations of high-altitude humidity measurements\cite{gierens2020well}) - result in large errors in contrail RF predictions from such models. Contrail detection work in aerial images can be largely attributed to Mannstein et al.\cite{mannstein1999operational}\cite{meyer2002regional}\cite{palikonda2005contrail}\cite{minnis2005contrail}\cite{meyer2007contrail}\cite{vazquez2010automatic}\cite{duda2013estimation}. This algorithm operates on brightness temperature imagery using a sequence of manually designed convolution and thresholding procedures, followed by recognizing appropriate linear linked components. The algorithm is tuned to have either high precision or recall; however, no single model has succeeded in achieving both simultaneously. Apart from this, deep learning models have been used for contrail detection using images of GOES-16 satellite\cite{kulik2019satellite} and images from Himawari-8 stationary satellite\cite{zhang2018contrail}. However, apart from \cite{weiss1998automatic} that has used Hough Transforms to segment AVHRR imagery, no work to our knowledge, has been done in over two decades to implement neural network-based segmentation models for contrails detection and Landsat-8 data.
%Semantic segmentation of satellite imagery of various regions, which can be further used to classify regions of high contrail density, is necessary to identify regions and implement this measure effectively. 

This work explores the potential of 'deep convolutional nets' to identify and segment contrails in satellite imagery.
For this purpose, we use popular semantic segmentation models, such as UNet\cite{ronneberger2015u}, PSP Net\cite{zhao2017pyramid}, DeepLab V3\cite{chen2017rethinking} and DeepLab V3+\cite{chen2018encoder} with different combinations of loss functions and encoders to adjust it accordingly for capturing the intricacies of contrails and achieve a more generalized result. We train these models on the labeled Landsat-8 contrails dataset proposed by \cite{mccloskeyhuman}, which consists of several scenes from the Landsat-8 satellite in which the contrails have been identified and marked as bounding polygons. We hope our proposed methods can be used as a benchmark for further work on this dataset and open the way for more substantial research on this domain.
\section{Data}
%We use the dataset published in 'A human-labeled Landsat-8 contrails dataset'\cite{mccloskeyhuman}, which has 4289 scenes primarily from 2018 of the Landsat-8 satellite and out of which 47\% scenes have at least one contrail. While prediction work has already been done on the GOES-16 and other NOAA series satellite images, using images of a low-earth orbit satellite such as Landsat-8 is beneficial for contrails as its images have a high spatial resolution of 30m and 100m per pixel for cirrus and thermal infrared bands respectively. This makes it easier to distinguish contrails from natural cirrus clouds, which are made of ice crystals and look like long and thin white lines. Each scene has a true-color RGB image, a false color Image, and labeled contrail points for this purpose. We use the False Colour Images and the contrail labeleds by generating them as a ground truth mask for each image. False Colour Images are generated by extracting the brightness temperature difference or red channel between the 12 µm and 11 µm bands, the 1.37 µm cirrus cloud reflectance band, which is the green channel and is omitted for nighttime images to avoid confusion and the 12 µm brightness temperature blue channel. The contrails in these images appear as black linear clouds, making it easier for the models to differentiate between cirrus clouds and contrails than RGB images for segmentation. This, as a result, helps improve the accuracy of these segmentation models. 
The dataset\cite{mccloskeyhuman} includes Landsat-8 scenes (primarily from 2018 and inside the viewable extent of the GOES-16 satellite), which have been reviewed by human labellers taught to identify and mark the bounding polygon of each contrail in the scene. The dataset has 4289 scenes primarily from 2018 of the Landsat-8 satellite, out of which 47\% of scenes have at least one contrail. 

While prediction work has already been done on the GOES-16 and other NOAA series satellite images, using images of a low-earth orbit satellite such as Landsat-8 is beneficial for contrails as its images have a high spatial resolution of 30m and 100m per pixel for cirrus and thermal infrared bands respectively. This makes it easier to distinguish contrails from natural cirrus clouds, which are made of ice crystals and look like long and thin white lines. Each scene has a true-color RGB image, a false color image, and labeled contrail points for this purpose.  

We use the false colour images and the contrail labels by generating them as a ground truth mask for each image. False colour images are generated by extracting the brightness temperature difference or red channel between the 12 µm and 11 µm bands, the 1.37 µm cirrus cloud reflectance band, which is the green channel and is omitted for nighttime images to avoid confusion and the 12 µm brightness temperature blue channel. The contrails in these images appear as black linear clouds, making it easier for the models to differentiate between cirrus clouds and contrails than RGB images for segmentation.

\section{Methodology}
% We approach our task as a binary Semantic Segmentation problem. After manually preprocessing and downloading images and masks after which they are trained on a variety of state of the art segmentation models with different combinations of encoders, encoder weights and losses to get the best combination of a segmentation model for contrails.

\subsection{Dataset Preprocessing}
The dataset is available in JSON string format stored in Google Cloud Storage. There are 100 files, each with the Landsat-8 filename, polygon bounds of contrails in the scene, and deidentified advected flight waypoints for each labeled scene. We manually preprocessed the data by constructing the ground truth mask from the polygon bounds using matplotlib collections\cite{Hunter:2007}. 
%We plotted RGB, False Colour Images, and the corresponding mask and saved the plots to maintain uniformity in the image shapes for each item.
We then resized the images and the masks to 512x512 dimensions for training; anything less than 512 led to a deterioration in results. We use only images with at least one contrail for training due to class imbalance between contrails and background and to help the model generalize better. The dataset has been further divided into 80\% for training (1737) and 20\% for testing (434). We opted to use the false color images as it was visually easier to identify the contrail than RGB images. Experimentally as well, the models performed better on false color images.

\subsection{Models}
We experimented with different types of state-of-the-art segmentation networks, namely UNet\cite{ronneberger2015u}, an encoder-decoder-based standard network, PSP Net\cite{zhao2017pyramid}, a pyramid pooling module-based network, DeepLabV3\cite{chen2017rethinking} and DeepLabV3+\cite{chen2018encoder} which are a combination of both encoder-decoder and pyramid pooling-based module. 
We used multiple backbones for each of these architectures. We employed Resnet101\cite{he2016deep},ResNeXt101-32x4d\cite{Xie2016} and Xception71\cite{chollet2017xception} with ImageNet pretrained weights.We experimented with both pretrained weights and random initialization and found that although results were almost the same, using pretrained weights helped the models to converge at least 30\% faster.
We chose these backbones for our segmentation models as\cite{zhang2020comparison} showed them as most effective for semantic segmentation of fine lines. 

Adam optimizer was used for all these instances with a learning rate of 0.0001 and sigmoid activation function. We tried various loss functions such as Focal\cite{https://doi.org/10.48550/arxiv.1708.02002}, Tversky\cite{https://doi.org/10.48550/arxiv.1706.05721}, Focal-Tversky\cite{https://doi.org/10.48550/arxiv.1810.07842}, Dice\cite{sudre2017generalised}, and Jaccard loss\cite{duque2021power}.  Intersection over Union (IoU) was used as the primary evaluation metric. We used different loss functions and noticed that an over-suppression of the Focal-Tversky is observed when the class accuracy is high, as the model is close to convergence. In contrast, for Dice loss, we noticed that it is very unstable as the model goes closer to converges and hence doesn't converge well.
To combat these shortcomings, we used a combination of Dice and Focal-Tversky loss, as shown in equation. \ref{eq:1}.
\begin{equation}\label{eq:1}
\centering
\begin{split}
TotalLoss = & \sum_c\Bigg(\delta\left(1-\frac{\sum_{i=1}^Np_{ic}g_{ic}+\epsilon}{\sum_{i=1}^Np_{ic}+g_{ic}+\epsilon}\right)+ \\ & (1-\delta)\left(1-\frac{\sum_{i=1}^Np_{ic}g_{ic}+\epsilon}{\sum_{i=1}^Np_{ic}g_{ic}+\alpha\sum_{i=1}^Np_{i\bar{c}}g_{ic}+\beta\sum_{i=1}^Np_{ic}g_{i\bar{c}}+\epsilon}\right)^{1/\gamma}\Bigg)
\end{split}
\end{equation}

N denotes the total number of pixels in an image, $g_{ic}$ and $p_{ic}$ represent the per pixel ground truth and predicted probability respectively for contrail class c, similarly $g_{i\bar{c}}$ and $p_{i\bar{c}}$ represent the non- contrail class $\bar{c}$. $\alpha$, $\beta$, and $\gamma$ are hyper-parameters for Focal-Tversky loss that can be tuned. $\delta$ is a hyper-parameter that decides the percentage of contribution of both Focal-Tversky and Dice loss towards the final loss calculated. After careful experimentation, we observed the best results with $\delta=0.5$.

\begin{figure}[ht]
\centering
\caption{Comparison of RGB, false colour, ground truth, and predicted mask for test images}
\includegraphics[width=1\textwidth]{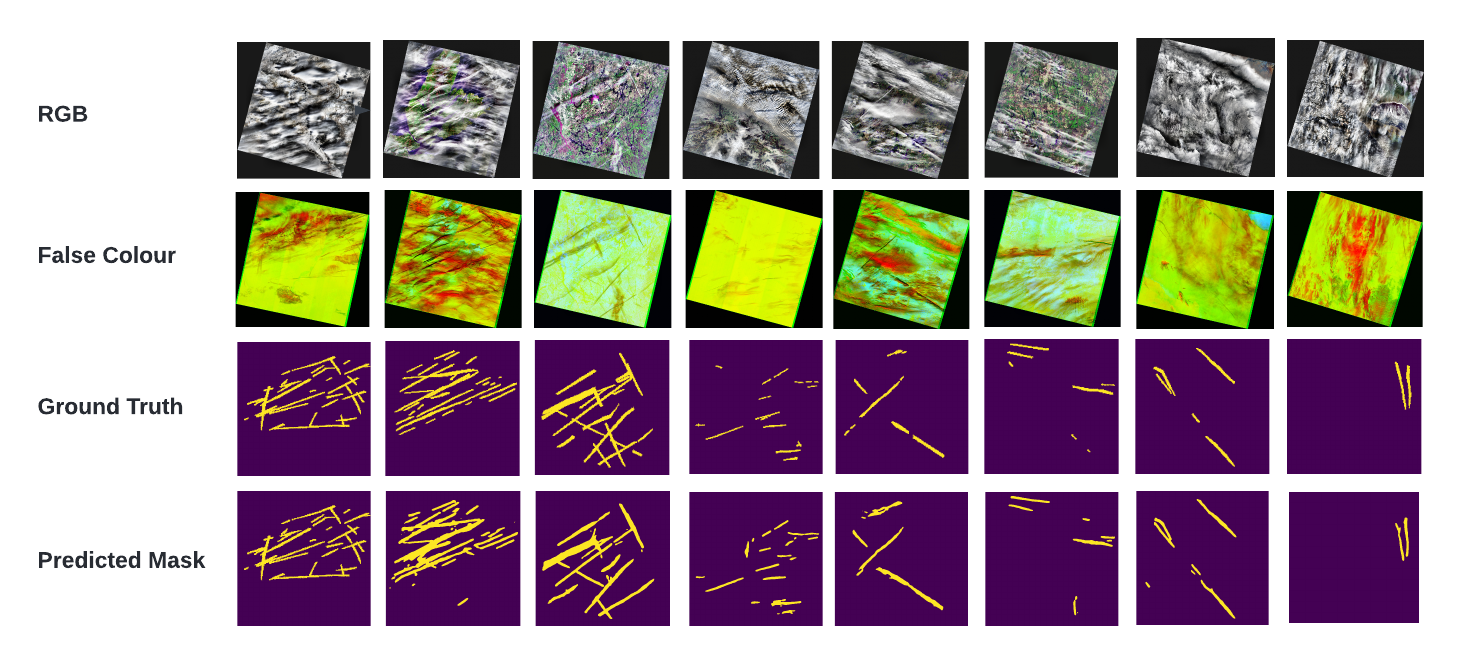}
\label{fig:contrail}
\end{figure}

\section{Results}
Table \ref{table:iou} compares the train and test IoU of all the segmentation models-encoder combinations. While there is about ±0.03 difference in the IoUs among all the combinations, we notice that the combination of UNet\cite{ronneberger2015u} architecture with Xception 71\cite{chollet2017xception} backbone gives the best IoU of \textbf{0.4395}. While the IoU was low, the model produced satisfactory contrail masks. We attribute this to the fact that the fundamental shape of the labels, i.e. thin and long hence IoU calculation is affected drastically if the prediction is even a few pixels off. We also tried simple data augmentation techniques, such as rotation and flipping, which didn't improve the IoU. Also, since we used false color images for our purpose, using hue saturation augmentation caused deterioration in our results.

We also noticed several mislabeled masks of contrails in the dataset. The dataset has several noisy labels due to manual labelling, due to which it suffers from large intra and inter-observer variability. Furthermore,  numerous contrail labels in the dataset are broader than the corresponding visible contrail in the original image. This and a severe class imbalance in the dataset were the two significant issues limiting the model's ability. We have used Dice and Focal-Tversky to create our loss function, as these are well known for dealing with class imbalance issues. We chose to consider images with at least one contrail for training. While these factors may affect the IoU, our combinations still have performed convincingly on the test dataset, evident in examples of the model's predictions against the ground truth, as provided in Figure \ref{fig:contrail}

\begin{table}\label{table}
\caption{IoU scores of segmentation models with different backbones}
  \label{sample-table}
  \centering
\begin{tabular}{
  l % left aligned column
  l % left aligned column
 *{4}{l}%{s[table-format=4.0]} % three columns with numeric data 
}
\toprule
  &  &\textbf{UNet} & \textbf{PSP Net} & \textbf{DeepLabV3} & \textbf{DeepLabV3+}\\
\midrule

\textbf{ResNet 101}  & Training &  0.6479 & 0.5032 &  0.4714 &  0.7048\\
    & Test   & 0.3410 & 0.3788  &  0.4143 &  0.4015\\
\textbf{ResNext 101-32x4d} & Training &  0.5411 & 0.6500 &  0.7211 &  0.6657\\
     & Test     &  0.4224 & \textbf{0.4044} &  \textbf{0.4339} &  \textbf{0.4266}\\
\textbf{Xception 71}    & Training &  0.6887 & 0.7272 &  0.7730 &  0.6290\\
     & Test     & \textbf{0.4395} & 0.4027 & 0.4246 &  0.4230\\
\bottomrule
\end{tabular}
\label{table:iou}
\end{table}
\section{Conclusion}
\label{gen_inst}
Contrails account for about 57\% of the global warming caused by the aviation industry\cite{LEE2021117834}. 
Robust contrail detection and segmentation is the first step in quantifying its lifetime impact and linking it to possible flights that may have caused it, thus assisting in tracking and avoiding contrails.

To the best of our knowledge, we are the first to present a detailed work on Landsat-8 data and evaluation on benchmarking of different state-of-the-art models for semantic segmentation for contrail detection.
Using false color images of Landsat-8 imagery, we achieve 0.4395 testing IoU UNet architecture with Xception 71 backbone. While the IoU is low, it fails to capture the model's true performance on the task. In our experiments, the model is more often than not able to segment contrails proficiently.

While this work is a solid first step in utilizing deep learning methods for contrail detection, there is much room for improvement. In future studies, we would like to experiment with temperature-based colour preprocessing. To test the model's robustness, we want to calculate uncertainty estimates under varying domains of satellite imagery and evaluate the model using a more suitable metric instead of IoU. Contextual information can also be incorporated based on publicly available datasets, such as
overlapping segmentation masks containing land cover, naturally occurring clouds, flight paths etc. Another avenue for future experiments could be using self-supervised and pseudo-labelling training techniques, attention-based
models, introduction of discriminator and vision transformers.

%Our future work involves using different low orbit satellite images to curate a more extensive and precise dataset. Additionally, transformer-based models can also be used for segmentation purposes. This can further be applied to an automatic segmentation and detection mechanism to understand the impact of contrails and air traffic in various regions. 
\section{Acknowledgments}
We would like to thank Mars Rover Manipal, an interdisciplinary student project team of MAHE, for providing the necessary resources for our research.
We are grateful to our faculty advisor, Dr Ujjwal Verma, for providing the necessary guidance.

\bibliography{tackling_climate_workshop.bib}

\begin{thebibliography}{10}

\bibitem{AVILA2019100033}
Denis Avila, Lance Sherry, and Terry Thompson.
\newblock Reducing global warming by airline contrail avoidance: A case study
  of annual benefits for the contiguous united states.
\newblock {\em Transportation Research Interdisciplinary Perspectives},
  2:100033, 2019.

\bibitem{mccloskeyhuman}
Kevin McCloskey, Scott Geraedts, Christopher Van~Arsdale, and Erica Brand.
\newblock A human-labeled landsat-8 contrails dataset.
\newblock In {\em ICML 2021 Workshop on Tackling Climate Change with Machine
  Learning}, 2021.

\bibitem{vazquez2015contrail}
Mannstein V{\'a}zquez-Navarro, Hermann Mannstein, and Stephan Kox.
\newblock Contrail life cycle and properties from 1 year of msg/seviri
  rapid-scan images.
\newblock {\em Atmospheric Chemistry and Physics}, 15(15):8739--8749, 2015.

\bibitem{bock2016reassessing}
Lisa Bock and Ulrike Burkhardt.
\newblock Reassessing properties and radiative forcing of contrail cirrus using
  a climate model.
\newblock {\em Journal of Geophysical Research: Atmospheres},
  121(16):9717--9736, 2016.

\bibitem{burkhardt2011global}
Ulrike Burkhardt and Bernd K{\"a}rcher.
\newblock Global radiative forcing from contrail cirrus.
\newblock {\em Nature climate change}, 1(1):54--58, 2011.

\bibitem{pearce2019airplane}
Fred Pearce.
\newblock How airplane contrails are helping make the planet warmer.
\newblock {\em YaleEnvironment360}, 2019.

\bibitem{molloy2022design}
Jarlath Molloy, Roger Teoh, Se{\'a}n Harty, George Koudis, Ulrich Schumann, Ian
  Poll, and Marc~EJ Stettler.
\newblock Design principles for a contrail-minimizing trial in the north
  atlantic.
\newblock {\em Aerospace}, 9(7):375, 2022.

\bibitem{mannstein2005note}
Hermann Mannstein, Peter Spichtinger, and Klaus Gierens.
\newblock A note on how to avoid contrail cirrus.
\newblock {\em Transportation Research Part D: Transport and Environment},
  10(5):421--426, 2005.

\bibitem{schumann2012contrail}
Ulrich Schumann.
\newblock A contrail cirrus prediction model.
\newblock {\em Geoscientific Model Development}, 5(3):543--580, 2012.

\bibitem{fritz2020role}
Thibaud~M Fritz, Sebastian~D Eastham, Raymond~L Speth, and Steven~RH Barrett.
\newblock The role of plume-scale processes in long-term impacts of aircraft
  emissions.
\newblock {\em Atmospheric Chemistry and Physics}, 20(9):5697--5727, 2020.

\bibitem{gierens2020well}
Klaus Gierens, Sigrun Matthes, and Susanne Rohs.
\newblock How well can persistent contrails be predicted?
\newblock {\em Aerospace}, 7(12):169, 2020.

\bibitem{mannstein1999operational}
Hermann Mannstein, Richard Meyer, and Peter Wendling.
\newblock Operational detection of contrails from noaa-avhrr-data.
\newblock {\em International Journal of Remote Sensing}, 20(8):1641--1660,
  1999.

\bibitem{meyer2002regional}
Richard Meyer, H~Mannstein, R~Meerk{\"o}tter, U~Schumann, and P~Wendling.
\newblock Regional radiative forcing by line-shaped contrails derived from
  satellite data.
\newblock {\em Journal of Geophysical Research: Atmospheres}, 107(D10):ACL--17,
  2002.

\bibitem{palikonda2005contrail}
Rabindra Palikonda, Patrick Minnis, David~P Duda, and Hermann Mannstein.
\newblock Contrail coverage derived from 2001 avhrr data over the continental
  united states of america and surrounding areas.
\newblock {\em Meteorologische Zeitschrift}, 14(4):525--536, 2005.

\bibitem{minnis2005contrail}
Patrick Minnis, Rabindra Palikonda, Bryan~J Walter, J~Kirk Ayers, and Hermann
  Mannstein.
\newblock Contrail coverage over the north pacific from avhrr data.
\newblock {\em Meteor. Z}, 2005.

\bibitem{meyer2007contrail}
Richard Meyer, R{\"u}diger Buell, Christian Leiter, Hermann Mannstein, Susanna
  Pechtl, Taikan Oki, and Peter Wendling.
\newblock Contrail observations over southern and eastern asia in noaa/avhrr
  data and comparisons to contrail simulations in a gcm.
\newblock {\em International Journal of Remote Sensing}, 28(9):2049--2069,
  2007.

\bibitem{vazquez2010automatic}
Mannstein Vazquez-Navarro, H~Mannstein, and B~Mayer.
\newblock An automatic contrail tracking algorithm.
\newblock {\em Atmospheric Measurement Techniques}, 3(4):1089--1101, 2010.

\bibitem{duda2013estimation}
David~P Duda, Patrick Minnis, Konstantin Khlopenkov, Thad~L Chee, and Robyn
  Boeke.
\newblock Estimation of 2006 northern hemisphere contrail coverage using modis
  data.
\newblock {\em Geophysical Research Letters}, 40(3):612--617, 2013.

\bibitem{kulik2019satellite}
Luke Kulik.
\newblock {\em Satellite-based detection of contrails using deep learning}.
\newblock PhD thesis, Massachusetts Institute of Technology, 2019.

\bibitem{zhang2018contrail}
Guoyu Zhang, Jinglin Zhang, and Jian Shang.
\newblock Contrail recognition with convolutional neural network and contrail
  parameterizations evaluation.
\newblock {\em SOLA}, 14:132--137, 2018.

\bibitem{weiss1998automatic}
John~M Weiss, Sundar~A Christopher, and Ronald~M Welch.
\newblock Automatic contrail detection and segmentation.
\newblock {\em IEEE transactions on geoscience and remote sensing},
  36(5):1609--1619, 1998.

\bibitem{ronneberger2015u}
Olaf Ronneberger, Philipp Fischer, and Thomas Brox.
\newblock U-net: Convolutional networks for biomedical image segmentation.
\newblock In {\em International Conference on Medical image computing and
  computer-assisted intervention}, pages 234--241. Springer, 2015.

\bibitem{zhao2017pyramid}
Hengshuang Zhao, Jianping Shi, Xiaojuan Qi, Xiaogang Wang, and Jiaya Jia.
\newblock Pyramid scene parsing network.
\newblock In {\em Proceedings of the IEEE conference on computer vision and
  pattern recognition}, pages 2881--2890, 2017.

\bibitem{chen2017rethinking}
Liang-Chieh Chen, George Papandreou, Florian Schroff, and Hartwig Adam.
\newblock Rethinking atrous convolution for semantic image segmentation.
\newblock {\em arXiv preprint arXiv:1706.05587}, 2017.

\bibitem{chen2018encoder}
Liang-Chieh Chen, Yukun Zhu, George Papandreou, Florian Schroff, and Hartwig
  Adam.
\newblock Encoder-decoder with atrous separable convolution for semantic image
  segmentation.
\newblock In {\em Proceedings of the European conference on computer vision
  (ECCV)}, pages 801--818, 2018.

\bibitem{Hunter:2007}
J.~D. Hunter.
\newblock Matplotlib: A 2d graphics environment.
\newblock {\em Computing in Science \& Engineering}, 9(3):90--95, 2007.

\bibitem{he2016deep}
Kaiming He, Xiangyu Zhang, Shaoqing Ren, and Jian Sun.
\newblock Deep residual learning for image recognition.
\newblock In {\em Proceedings of the IEEE conference on computer vision and
  pattern recognition}, pages 770--778, 2016.

\bibitem{Xie2016}
Saining Xie, Ross Girshick, Piotr Dollár, Zhuowen Tu, and Kaiming He.
\newblock Aggregated residual transformations for deep neural networks.
\newblock {\em arXiv preprint arXiv:1611.05431}, 2016.

\bibitem{chollet2017xception}
Fran{\c{c}}ois Chollet.
\newblock Xception: Deep learning with depthwise separable convolutions.
\newblock In {\em Proceedings of the IEEE conference on computer vision and
  pattern recognition}, pages 1251--1258, 2017.

\bibitem{zhang2020comparison}
Rongyu Zhang, Lixuan Du, Qi~Xiao, and Jiaming Liu.
\newblock Comparison of backbones for semantic segmentation network.
\newblock In {\em Journal of Physics: Conference Series}, volume 1544, page
  012196. IOP Publishing, 2020.

\bibitem{https://doi.org/10.48550/arxiv.1708.02002}
Tsung-Yi Lin, Priya Goyal, Ross Girshick, Kaiming He, and Piotr Dollár.
\newblock Focal loss for dense object detection, 2017.

\bibitem{https://doi.org/10.48550/arxiv.1706.05721}
Seyed Sadegh~Mohseni Salehi, Deniz Erdogmus, and Ali Gholipour.
\newblock Tversky loss function for image segmentation using 3d fully
  convolutional deep networks, 2017.

\bibitem{https://doi.org/10.48550/arxiv.1810.07842}
Nabila Abraham and Naimul~Mefraz Khan.
\newblock A novel focal tversky loss function with improved attention u-net for
  lesion segmentation, 2018.

\bibitem{sudre2017generalised}
Carole~H Sudre, Wenqi Li, Tom Vercauteren, Sebastien Ourselin, and
  M~Jorge~Cardoso.
\newblock Generalised dice overlap as a deep learning loss function for highly
  unbalanced segmentations.
\newblock In {\em Deep learning in medical image analysis and multimodal
  learning for clinical decision support}, pages 240--248. Springer, 2017.

\bibitem{duque2021power}
David Duque-Arias, Santiago Velasco-Forero, Jean-Emmanuel Deschaud, Francois
  Goulette, Andr{\'e}s Serna, Etienne Decenci{\`e}re, and Beatriz Marcotegui.
\newblock On power jaccard losses for semantic segmentation.
\newblock In {\em VISAPP 2021: 16th International Conference on Computer Vision
  Theory and Applications}, 2021.

\bibitem{LEE2021117834}
D.-S. Lee, David~W. Fahey, Agnieszka Skowron, Myles~R. Allen, Ulrike Burkhardt,
  Qi~Chen, Sarah~J. Doherty, Sarah Freeman, Piers~M. Forster, Jan~S.
  Fuglestvedt, Andrew Gettelman, Rub{\'e}n Rodr{\'i}guez~De Le{\'o}n, Ling~L.
  Lim, Marianne~Tronstad Lund, Richard~J. Millar, Bethan Owen, J.~E. Penner,
  Giovanni~Mario Pitari, Michael~J. Prather, Robert Sausen, and Laura~J.
  Wilcox.
\newblock The contribution of global aviation to anthropogenic climate forcing
  for 2000 to 2018.
\newblock {\em Atmospheric Environment (Oxford, England : 1994)}, 244:117834 --
  117834, 2020.

\end{thebibliography}
\end{document}